\documentclass[10pt,twocolumn,letterpaper]{article}

\usepackage[pagenumbers]{cvpr} 

\usepackage{booktabs}
\definecolor{cvprblue}{rgb}{0.21,0.49,0.74}
\usepackage[pagebackref,breaklinks,colorlinks,allcolors=cvprblue]{hyperref}
\usepackage{tabularx}
\usepackage{arydshln}

\def\paperID{9} 
\def\confName{CVPR}
\def\confYear{2025}

\title{EgoVIS@CVPR: PAIR-Net: Enhancing Egocentric Speaker Detection via Pretrained Audio–Visual Fusion and Alignment Loss}

\author{
Yu Wang \quad Juhyung Ha \quad David J. Crandall\\
Indiana University Bloomington\\
Bloomington, IN, USA\\
{\tt\small \{yw173, juhha, djcran\}@iu.edu}
}

\begin{document}
\maketitle
\begin{abstract}
Active speaker detection (ASD) in egocentric videos presents unique challenges due to unstable viewpoints, motion blur, and off-screen speech sources—conditions under which traditional visual-centric methods degrade significantly. We introduce PAIR-Net (\textbf{P}retrained \textbf{A}udio–visual \textbf{I}ntegration with \textbf{R}egularization \textbf{Net}work), a  model that integrates a partially frozen Whisper audio encoder with a fine-tuned AV-HuBERT visual backbone to robustly fuse cross-modal cues. To counteract modality imbalance, we introduce an inter-modal alignment loss that synchronizes audio and visual representations, enabling more consistent convergence across modalities. Without relying on multi-speaker context or ideal frontal views, PAIR-Net achieves state-of-the-art performance on the Ego4D ASD benchmark with 76.6\% mAP, surpassing LoCoNet and STHG by 8.2\% and 12.9\% mAP, respectively. Our results highlight the value of pretrained audio priors and alignment-based fusion for robust ASD under real-world egocentric conditions.
\end{abstract}
    
\section{Introduction}
\label{sec:intro}
Active speaker detection (ASD) is a fundamental task in multimodal understanding, supporting applications such as speaker diarization~\cite{qiao2024joint, chung2019said}, speech understanding~\cite{chung2018learning}, and human–robot interaction~\cite{skowronski2023active}. While prior methods~\cite{tao2021someone, liao2023light} perform well on third-person, high-quality footage with clearly visible faces, significant challenges remain in unconstrained settings. Specifically, egocentric videos from head-mounted cameras often suffer from motion blur, oblique angles, and low-resolution faces, while speech may originate from off-screen speakers.

To address these challenges, recent methods such as LoCoNet~\cite{wang2024loconet} and STHG~\cite{min2023sthg} have aimed to improve visual modeling by incorporating multi-speaker contextual reasoning. While these approaches enhance visual robustness, they often underutilize the rich information embedded in the audio modality, limiting the full exploitation of cross-modal complementarities.

In this work, we propose the \textbf{P}retrained \textbf{A}udio–visual \textbf{I}ntegration with \textbf{R}egularization \textbf{Net}work (PAIR-Net), which integrates an audio encoder from Whisper~\cite{radford2023robust} and a video encoder from AV-HuBERT~\cite{shi2022learning}, followed by a lightweight one-layer decoder. Whisper~\cite{radford2023robust}, an audio-based speech recognition model trained on large-scale audio data in a weakly supervised manner, provides strong priors on human speech. AV-HuBERT~\cite{shi2022learning}, developed for audio-visual speech recognition and capable of performing lip reading using visual input alone, indicates that its video encoder enables robust modeling of lip dynamics. We leverage the pretrained weights of both encoders for transfer learning: Whisper’s~\cite{radford2023robust} audio encoder is partially frozen to retain its low-level speech representations, while the entire AV-HuBERT video encoder ~\cite{shi2022learning} is fine-tuned to adapt its lip reading priors to full-face sequences. This design encourages the model to exploit the complementary strengths of both modalities.

We observe that the model often over-relies on visual cues while underutilizing audio, a phenomenon we term \textit{modality imbalance}. To address this, we introduce an inter-modal alignment loss—motivated by LFM~\cite{yang2024facilitating}—that enforces agreement between temporally aligned audio and visual features. This regularization balances  across modalities and, when combined with the standard classification loss, enhances overall performance. PAIR-Net achieves strong results using input from a single visible speaker, demonstrating both efficiency and generalization. On the Ego4D ASD benchmark \cite{grauman2022ego4d}, we  attain 76.6\% mAP, surpassing prior state-of-the-art models LoCoNet~\cite{wang2024loconet} and STHG~\cite{min2023sthg} by 8.2\% and 12.9\%, respectively.

Our contributions are summarized as follows:
\begin{itemize}
    \item We propose PAIR-Net, a novel audio–visual alignment network that integrates pretrained Whisper~\cite{radford2023robust} and AV-HuBERT~\cite{shi2022learning} encoders for robust active speaker detection in egocentric scenarios.
    \item We introduce an inter-modal alignment loss to the ASD task, encouraging convergence consistency between audio and visual features during training, which effectively mitigates modality imbalance.
    \item We achieve state-of-the-art results on the Ego4D ASD benchmark~\cite{grauman2022ego4d}, demonstrating the model’s efficiency and generalization using only a single visible speaker.
\end{itemize}

\section{Related Work}
\label{sec:formatting}
ASD methods like TalkNet~\cite{tao2021someone} and LightASD~\cite{liao2023light} capture temporal features but rely on clearly visible faces, limiting their robustness in egocentric videos with motion blur and oblique angles. Recent context-aware models such as LoCoNet~\cite{wang2024loconet}, SPELL~\cite{min2022intel, min2022learning}, and STHG~\cite{min2023sthg} improve performance by modeling speaker interactions or spatio-temporal relations, yet they assume access to rich multi-speaker context and high-quality input. To enhance robustness, ASDnB~\cite{roxo2024asdnb} adds body cues, while LASER~\cite{nguyen2025laser} incorporates lip landmarks. In contrast, PAIR-Net is designed for egocentric scenarios with limited visual information from a single-face video stream, overcoming the limitations of prior context- and profile-dependent models.

Multi-modal models often exhibit \textit{modality imbalance}, where one modality dominates the learning process, which may result in \textit{modality laziness}~\cite{zhang2024multimodal, du2023uni}. G-Blend~\cite{wang2020makes} addresses the imbalance by dynamically reweighting gradients according to each modality’s overfitting behavior. Greedy~\cite{wu2022characterizing} attributes suboptimal convergence to the early dominance of easier modalities, motivating curriculum-based scheduling. AGM~\cite{li2023boosting} and PMR~\cite{fan2023pmr} introduce customized learning strategies for different modalities. Our model is motivated by LFM~\cite{yang2024facilitating}, which demonstrates that introducing a alignment loss helps alleviate inconsistent label fitting across modalities. Building on this, we freeze the early layers of Whisper~\cite{radford2023robust} to retain its pretrained audio priors and incorporate an inter-modal alignment loss to promote balanced audio-visual learning.

\section{Methodology}

\begin{figure}[t]
  \centering
   \includegraphics[width=1\linewidth]{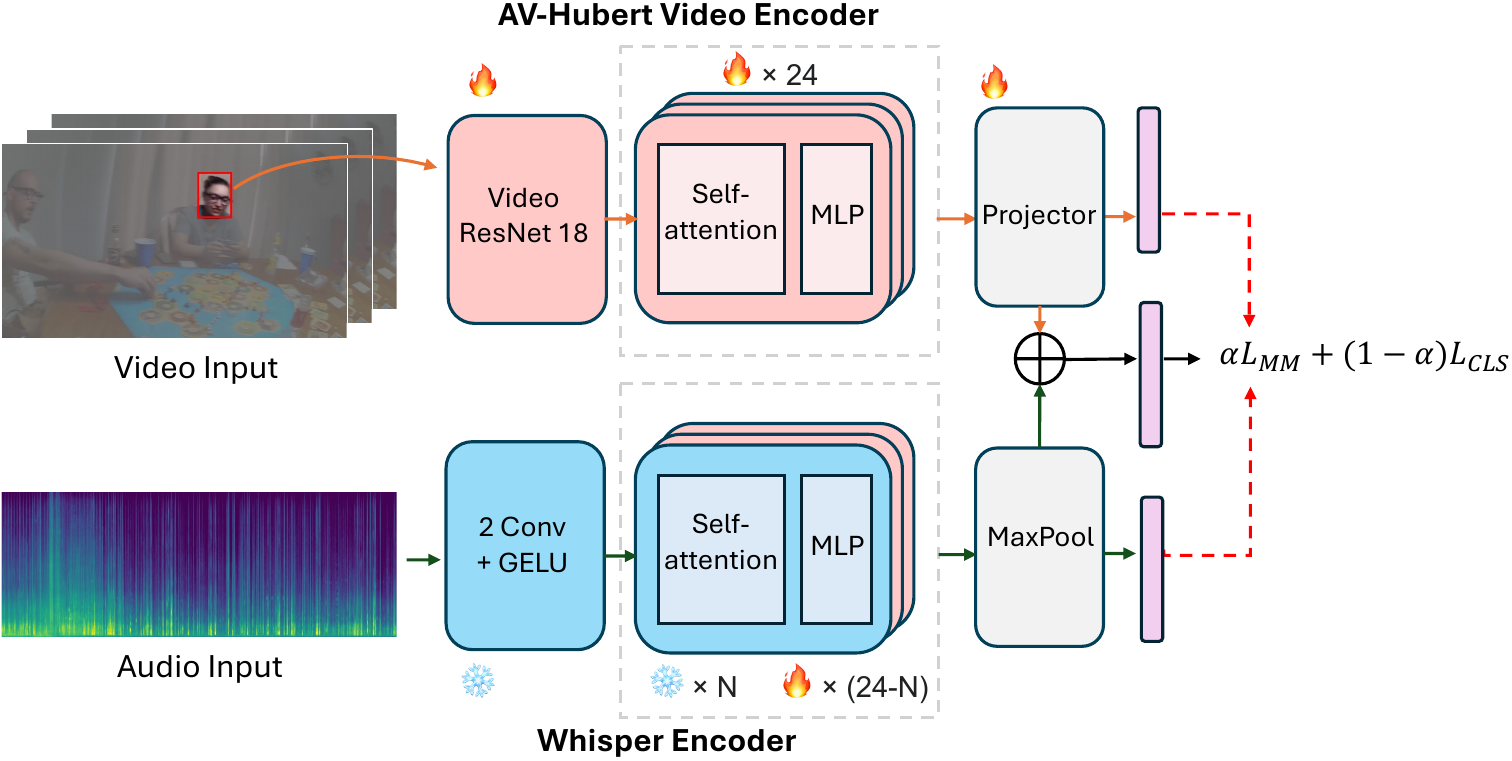}

   \caption{An overview of  PAIR-Net. Modality-specific encoders process audio and visual inputs, with the early layers of the audio encoder frozen during training. The video encoder is followed by a trainable projector, while the audio encoder includes a temporal max pooling layer to align sequence lengths. A fully connected layer predicts the ASD label using the fused representation. Two additional modality-specific classifiers generate individual predictions, which are used to compute the inter-modal alignment loss.}

   \label{fig:overall}
\end{figure}

We propose PAIR-Net, an audio-visual active speaker detection (ASD) framework that combines powerful pretrained encoders with a simple yet effective fusion mechanism and contrastive supervision to mitigate modality imbalance.
%
%
Given a sequence of video frames $I_v \in \mathbb{R}^{T_v \times H \times W}$ containing cropped face regions and the corresponding log-Mel spectrogram $I_a \in \mathbb{R}^{T_a \times 80}$ derived from the raw audio, the goal is to predict frame-wise speaker activity.

\subsection{Model Architecture}

The PAIR-Net architecture is illustrated in Figure~\ref{fig:overall}. We use the AV-HuBERT video encoder~\cite{shi2022learning} and the Whisper~\cite{radford2023robust} audio encoder for robust multimodal feature extraction. The video encoder is based on the AV-HuBERT visual branch~\cite{shi2022learning}, which consists of a ResNet backbone \cite{he2016deep} for low-level feature extraction and 24 Transformer blocks for capturing high-level spatiotemporal features. For audio, we use the Whisper encoder~\cite{radford2023robust}, composed of two convolutional layers followed by 24 Transformer blocks, yielding temporally rich representations.

The video encoder is followed by a projection head, implemented as a fully-connected layer, to transform video features. Due to the STFT-based preprocessing, the audio sequence is temporally longer than the corresponding video sequence. To enable temporal-aligned fusion, we apply an adaptive max pooling operation to the audio feature sequence, aligning it to the video temporal length $T_v$.

Let $z_{(i)} = \phi_{(i)}(I_{(i)}; \Theta_{(i)})$ be the modality-specific features for $i \in \{v, a\}$, where $z_{(i)} \in \mathbb{R}^{T_{(i)} \times f}$ and $\Theta_{(i)}$ denotes the parameters of the $i$-th encoder. The fusion pipeline is defined as:
\begin{align}
    z_v' &= \text{Proj}(z_v), \\
    z_a' &= \text{MaxPool}(z_a; T_v), \\
    y &= \text{softmax}(W(z_v' + z_a') + b),
\end{align}
where $W$ and $b$ denote the parameters of the final classification layer.

Despite its simplicity, this additive fusion strategy achieves comparable performance with the cross-attention variants in our experiments, highlighting the strength of pretrained encoders in generating highly discriminative unimodal features.

The primary classification objective is defined as:
\begin{equation}
    \mathcal{L}_{\text{CLS}} = -\frac{1}{n} \sum_{j=1}^{n} \hat{y}^{(j)} \log y^{(j)},
\end{equation}
where $\hat{y}^{(j)}$ is the one-hot ground truth label for frame $j$.

\subsection{Modality Alignment}

While visual cues dominate ASD performance, we observe suboptimal utilization of the audio modality during joint training. To address this imbalance, we draw inspiration from LFM~\cite{yang2024facilitating} and introduce an alignment loss that encourages alignment between the label fitting across modalities.

To preserve the pretrained knowledge of Whisper~\cite{radford2023robust}, we freeze its initial layers (including the 1D-convolutional layers and the first $N$ Transformer blocks), allowing only the later layers to adapt. We then apply classification heads independently to each modality:
\begin{align}
    y_v &= \text{softmax}(W_v z_v' + b_v), \\
    y_a &= \text{softmax}(W_a z_a' + b_a), \\
    \tilde{y} &= \frac{1}{2}(y_v+y_a),
\end{align}
and then compute the inter-modal alignment loss to align the label fitting of each modality:
\begin{equation}
    \mathcal{L}_{\text{MM}} = \frac{1}{2} \text{KL}(y_v \| \tilde{y}) + \frac{1}{2} \text{KL}(y_a \| \tilde{y}).
\end{equation}

The final training objective combines the classification loss and the alignment loss:
\begin{equation}
    \mathcal{L} = (1 - \alpha) \mathcal{L}_{\text{CLS}} + \alpha \mathcal{L}_{\text{MM}},
\end{equation}
where $\alpha \in [0, 1]$ balances the two components.

\section{Experiments}
We evaluate  on the Ego4D Audio-Visual Benchmark~\cite{grauman2022ego4d}, focusing on Active Speaker Detection (ASD).

\subsection{Datasets}
The Ego4D dataset~\cite{grauman2022ego4d} is a large-scale collection of egocentric videos. In this work, we utilize the audio-visual benchmark subset, which comprises 3,670 hours of first-person video footage. The egocentric perspective introduces unique challenges, such as significant camera motion and frequent off-screen speech.
\subsection{Experimental Setting}
We extract face sequences from each frame using the provided bounding boxes, resize them to $112\times112$, and stack them to form the video input. Corresponding audio is converted to log‑Mel spectrograms via Whisper’s ~\cite{radford2023robust} preprocessing pipeline. Both the video and audio encoders are initialized with pretrained weights from the AV‑HuBERT and Whisper large‑v3 models~\cite{radford2023robust}, respectively. We freeze the first two convolutional layers and the first 10 Transformer blocks of the Whisper encoder~\cite{radford2023robust}. Following the LFM~\cite{yang2024facilitating} setup, the loss weighting coefficient $\alpha$ is initialized at approximately 0.8 and gradually decays during training. We train the PAIR-Net model using AdamW~\cite{loshchilov2017decoupled} with a batch size of 1,500 frames for four epochs on a single NVIDIA L40S GPU. The learning rate is set to $5\times10^{-5}$ and decays by 5\% each epoch.

\subsection{Comparison with the State-of-the-art}
We reported the performance of our PAIR-Net on the Ego4D ASD dataset~\cite{grauman2022ego4d} comparing with previous state-of-the-art models.

\begin{table}[t]
\centering
\small
\begin{tabular*}{0.9\columnwidth}{@{\extracolsep{\fill}}l |>{\centering\arraybackslash}p{3em}}
\toprule
\textbf{Method} & \textbf{mAP} \\
\midrule
TalkNet (MM'21)~\cite{tao2021someone}      & $51.7^\dagger$ \\
LightASD (CVPR'23)~\cite{liao2023light}     & $52.8$ \\
SPELL (ECCV'22)~\cite{min2022intel, min2022learning}        & $60.7^\dagger$ \\
STHG (CVPR Workshop'23)~\cite{min2023sthg}         & $63.7^\dagger$ \\
LoCoNet (CVPR'24)~\cite{wang2024loconet}      & $68.4^\dagger$ \\
\midrule
PAIR-Net & \textbf{76.6}\\
\bottomrule
\end{tabular*}
\vspace{-0.5em}
\caption{Comparison of mAP scores on Ego4D ASD benchmarks~\cite{grauman2022ego4d}. $^\dagger$ denotes that score is obtained from the papers.}
\label{table:asd}
\end{table}

As shown in Table~\ref{table:asd}, we adopt mean Average Precision (mAP) as our evaluation metric. Prior methods such as TalkNet~\cite{tao2021someone} and LightASD~\cite{liao2023light} achieve moderate performance but lack robustness in egocentric scenarios. Context-aware approaches, including SPELL~\cite{min2022intel}, STHG~\cite{min2023sthg}, and LoCoNet~\cite{wang2024loconet}, leverage multi-speaker cues or graph-based reasoning to yield stronger results; however, they rely on high-quality, multi-person context. In contrast, PAIR-Net operates on a single-face video stream and outperforms all baselines with an mAP of 76.6\%, demonstrating the effectiveness of our proposed framework.

\section{Conclusion}
We proposed PAIR-Net, an active speaker detection model that fuses pretrained Whisper~\cite{radford2023robust} and AV-HuBERT~\cite{shi2022learning} encoders with a prediction-level alignment loss. Selectively freezing the early layers of the audio encoder preserves speech priors, while the alignment objective mitigates modality imbalance. PAIR-Net achieves state-of-the-art results on the Ego4D ASD benchmark~\cite{grauman2022ego4d}, demonstrating strong robustness and generalization in egocentric settings. Our findings underscore the effectiveness of pretrained audio priors and inter-modal alignment for real-world egocentric multimodal understanding.

\section{Acknowledgments}
This
work was supported in part by the National Science Foundation under award DRL-2112635 to the AI Institute for Engaged Learning. Any opinions, findings,  conclusions,
or recommendations   are those of
the author(s) and do not necessarily reflect those  of the NSF.

{
    \small
    \bibliographystyle{ieeenat_fullname}
    \bibliography{main}
}


\end{document}